\documentclass[a4paper]{article}
\usepackage{RR}
\usepackage{hyperref}

\usepackage{epsfig}
\usepackage{algorithmic,algorithm}

\usepackage{amsthm}
\usepackage{amsfonts}
\usepackage{amssymb}
\usepackage{amsmath}

\newcommand{\bi}{\begin{itemize}}
\newcommand{\ei}{\end{itemize}}

\newcommand{\beq}{\begin{equation}}
\newcommand{\eeq}{\end{equation}}

\newcommand{\beqa}{\begin{eqnarray}}
\newcommand{\eeqa}{\end{eqnarray}}

\newcommand{\beqan}{\begin{eqnarray*}}
\newcommand{\eeqan}{\end{eqnarray*}}

\newcommand{\EE}{{\cal{E}}}
\newcommand{\PP}[1]{{\mathbb P}\left(#1\right)}
\newcommand{\E}{{\mathbb E}}

\renewcommand{\L}{{\cal{L}}}
\newcommand{\F}{{\cal{F}}}
\newcommand{\B}{{\cal{B}}}
\newcommand{\C}{{\cal{C}}}
\renewcommand{\S}{{\cal{S}}}

\newcommand{\FF}{{\mathcal F}}

\newcommand{\eqdef}{\stackrel{{\rm\scriptsize def}}{=}}

\newtheorem{theorem}{Theorem}

\newtheorem{prop}{Proposition}
\newtheorem{remark}{Remark}

\RRdate{March 2007}

\RRauthor{
Pierre-Arnaud Coquelin \thanks{CMAP, Ecole Polytechnique, \texttt{coquelin@cmapx.polytechnique.fr}}%
  \and
R\'emi Munos \thanks{SequeL, INRIA, \texttt{remi.munos@inria.fr}}
}
\authorhead{Coquelin \& Munos}
\RRtitle{Bandit Algorithms for Tree Search}
\RRetitle{Bandit Algorithms for Tree Search}
\titlehead{Bandit Algorithms for Tree Search}
\RRresume{Les m\'ethodes de recherche arborescentes utilisant des algorithmes de bandit ont r\'ecemment connu une forte popularit\'e, pour leur capacit\'e de traiter des grands arbres, par exemple pour le jeu de go \cite{Gelly06}. Il est connu que l'algorithme UCT \cite{Kocsis06}, une m\'ethode de recherche arborescente bas\'ees sur des intervalles de confiance (algorithme {\em Upper Confidence Bounds} (UCB) de \cite{Auer02}), s'adapte locallement \`a la profondeur effective de l'arbre. Cependant, nous montrons ici que UCT peut \^etre trop ``optimiste'' dans certains cas, menant \`a un regret $\Omega(\exp(\exp(D)))$ o\`u $D$ est la profondeur de l'arbre. Nous proposons plusieurs alternatives d'algorithmes de bandit pour la recherche arborescente. Tout d'abord, nous proposons une modification d'UCT utilisant un intervalle de confiance qui cro\^it exponentiellement avec la profondeur de l'horizon de l'arbre, et montrons qu'il m\`ene \`a un regret $O(2^D \sqrt{n})$ mais ne s'adapte pas \`a la r\'egularit\'e de l'arbre. Puis nous analysons un algorithme {\em Flat-UCB} de bandit de type UCB directement sur les feuilles et prouvons une borne finie (ind\'ependante de $n$) sur le regret avec forte probabilit\'e.
Ensuite, nous introduisons un algorithme {\em Bandit Algorithm for Smooth Trees} qui prend en compte d'\'eventuelles r\'egularit\'es dans l'arbre pour r\'ealiser des ``coupes'' efficaces de branches sous-optimale avec grande confiance. Enfin, nous pr\'esentons une version incr\'ementale de recherche arborescente qui s'applique lorsque l'arbre est trop grand (voire infini) pour pouvoir \^etre repr\'esent\'e enti\`erement, et montrons qu'essentiellement, et avec forte probabilit\'e, seule la branche optimale est ind\'efiniment d\'evelopp\'ee. Nous illustrons ces m\'ethodes sur un probl\`eme d'optimisation d'une fonction Lipschitzienne, \`a partir de donn\'ees bruit\'ees.}

\RRabstract{Bandit based methods for tree search have recently gained popularity when applied to huge trees, e.g. in the game of go \cite{Gelly06}. The {\em UCT algorithm} \cite{Kocsis06}, a tree search method based on Upper Confidence Bounds (UCB) \cite{Auer02}, is believed to adapt locally to the effective smoothness of the tree. However, we show that UCT is too ``optimistic'' in some cases, leading to a regret $\Omega(\exp(\exp(D)))$ where $D$ is the depth of the tree. We propose alternative bandit algorithms for tree search. First, a modification of UCT using a confidence sequence that scales exponentially with the horizon depth is proven to have a regret $O(2^D \sqrt{n})$, but does not adapt to possible smoothness in the tree. We then analyze {\em Flat-UCB} performed on the leaves and provide a finite regret bound with high probability. Then, we introduce a UCB-based {\em Bandit Algorithm for Smooth Trees} which takes into account actual smoothness of the rewards for performing efficient ``cuts'' of sub-optimal branches with high confidence. Finally, we present an incremental tree search version which applies when the full tree is too big (possibly infinite) to be entirely represented and show that with high probability, essentially only the optimal branches is indefinitely developed. We illustrate these methods on a global optimization problem of a Lipschitz function, given noisy data.}
\RRmotcle{Algorithmes de bandit, recherche arborescente, compromis exploration-exploitation, bornes sup\'erieures d'intervalles de confiance, jeux minimax, apprentissage par renforcement}
\RRkeyword{Bandit algorithms, tree search, exploration-exploitation tradeoff, upper confidence bounds, minimax game, reinforcement learning }
\RRprojet{SequeL}
\RRtheme{\THCog} 
\URFuturs 
\begin{document}
\makeRR   

\section{Introduction}
Bandit algorithms have been used recently for tree search, because of their efficient trade-off between exploration of the most uncertain branches and exploitation of the most promising ones, leading to very promising results in dealing with huge trees (e.g. the go program MoGo, see \cite{Gelly06}). In this paper we focus on Upper Confidence Bound (UCB) bandit algorithms  \cite{Auer02} applied to tree search, such as UCT (Upper Confidence Bounds applied to Trees) \cite{Kocsis06}. The general procedure is described by Algorithm \ref{alg:BATS} and depends on the way the upper-bounds $B_{i,p,n_i}$ for each node $i$ are maintained. 

\begin{algorithm}[!]
   \caption{Bandit Algorithm for Tree Search}
   \label{alg:BATS}
\begin{algorithmic}
   \FOR{$n\geq 1$}
   \STATE {\bf Run $n$-th trajectory from the root to a leaf:}
   \STATE Set the current node $i_0$ to the root
   \FOR{$d=1$ {\bfseries to} $D$}
   \STATE Select node $i_d$ as the children $j$ of node $i_{d-1}$ that maximizes $B_{j,n_{i_{d-1}},n_j}$
   \ENDFOR
   \STATE Receive reward $x_{n}\stackrel{iid}{\sim} X_{i_D}$
   \STATE {\bf Update the nodes visited by this trajectory:}
   \FOR{$d=D$ {\bfseries to} $0$}
   \STATE Update the number of visits: $n_{i_d}=n_{i_d}+1$
   \STATE Update the bound $B_{i_d, n_{i_{d-1}}, n_{i_d}}$
   \ENDFOR
   \ENDFOR
\end{algorithmic}
\end{algorithm}

A trajectory is a sequence of nodes from the root to a leaf, where at each node, the next node is chosen as the one maximizing its $B$ value among the children. A reward is received at the leaf. After a trajectory is run, the $B$ values of each node in the trajectory are updated.
In the case of UCT, the upper-bound $B_{i,p,n_i}$ of a node $i$, given that the node has already been visited $n_i$ times and its parent's node $p$ times, is the average of the rewards $\{x_t\}_{1\leq t\leq n_i}$ obtained from that node $X_{i,n_i}=\frac{1}{n_i} \sum_{t=1}^{n_i} x_t$ plus a confidence interval, derived from a Chernoff-Hoeffding bound (see e.g. \cite{Gyorfi96}):
\beq \label{eq:uct} 
B_{i,p,n_i} \eqdef X_{i,n_i} + \sqrt{\frac{2\log(p)}{n_i}}
\eeq

In this paper we consider a $\max$ search (the minimax problem is a direct generalization of the results presented here) in binary trees (i.e. there are $2$ actions in each node), although the extension to more actions is straightforward. Let a binary tree of depth $D$ where at each leaf $i$ is assigned a random variable $X_i$, with bounded support $[0,1]$, whose law is unknown. Successive visits of a leaf $i$ yield a sequence of independent and identically distributed (i.i.d.) samples $x_{i,t}\sim X_i$, called rewards, or payoff. The value of a leaf $i$ is its expected reward: $\mu_i\eqdef\E X_i$. Now we define the value of any node $i$ as the maximal value of the leaves in the branch starting from node $i$. Our goal is to compute the value $\mu^*$ of the root.

An optimal leaf is a leaf having the largest expected reward. We will denote by $*$ quantities related to an optimal node. For example $\mu^*$ denote $\max_i \mu_i$. An optimal branch is a sequence of nodes from the root to a leaf, having the $\mu^*$ value. We define the {\em regret} up to time $n$ as the difference between the optimal expected payoff and the sum of obtained rewards:
$$ R_n \eqdef \mu^* - \sum_{t=1}^{n} x_{i_t,t},$$
where $i_t$ is the chosen leaf at round $t$. We also define the {\em pseudo-regret} up to time $n$:
$$ \bar R_n \eqdef \mu^* - \sum_{t=1}^{n} \mu_{i_t} = \sum_{j\in\L} n_j \Delta_j,$$
where $\L$ is the set of leaves, $\Delta_j \eqdef \mu^*-\mu_j$, and $n_j$ is the random variable that counts the number of times leaf $j$ has been visited up to time $n$. The pseudo-regret may thus be analyzed by estimating the number of times each sub-optimal leaf is visited. 

In tree search, our goal is thus to find an exploration policy of the branches such as to minimize the regret, in order to select an optimal leaf as fast as possible. Now, thanks to a simple contraction of measure phenomenon, the regret per bound $R_n/n$ turns out to be very close to the pseudo regret per round $\bar R_n/n$. Indeed, using Azuma's inequality for martingale difference sequences (see Proposition \ref{prop:Azuma}), with probability at least $1-\beta$, we have at time $n$, $$ \frac 1n|R_n - \bar R_n| \leq \sqrt{\frac{2\log(2/\beta)}{n}}.$$ 

The fact that $R(n)-\bar R_n$ is a martingale difference sequence comes from the property that, given the filtration $\F_{t-1}$ defined by the random samples up to time $t-1$, the expectation of the next reward $\E x_t$ is conditioned to the leaf $i_t$ chosen by the algorithm: $\E[x_t|\F_{t-1}]=\mu_{i_t}$. Thus $R_n-\bar R_n = \sum_{t=1}^n x_t-\mu_{i_t}$ with $\E[x_t-\mu_{i_t}|\F_{t-1}]=0$.
Hence, we will only focus on providing high probability bounds on the pseudo-regret.

First, we analyze the UCT algorithm defined by the upper confidence bound (\ref{eq:uct}). We show that its behavior is risky and may lead to a regret as bad as $\Omega(\exp(\cdots\exp(D)\cdots))$ ($D-1$ composed exponential functions). We modify the algorithm by increasing the exploration sequence, defining:
\beq \label{eq:uct_sqrt} 
B_{i,p,n_i} \eqdef X_{i,n_i} + \sqrt{\frac{\sqrt{p}}{n_i}}.
\eeq

This yields an improved worst-case behavior over regular UCT, but the regret may still be as bad as $\Omega(\exp(\exp(D)))$ (see Section \ref{sec:UCT}).
We then propose in Section \ref{sec:modified.UCT} a {\em modified UCT} based on the bound (\ref{eq:uct_sqrt}), where the confidence interval is multiplied by a factor that scales exponentially with the horizon depth. We derive a worst-case regret $O(2^D/\sqrt{n})$ with high probability. However this algorithm does not adapt to the effective smoothness of the tree, if any.

Next we analyze the {\em Flat-UCB} algorithm, which simply performs UCB directly on the leaves. With a slight modification of the usual confidence sequence, we show in Section \ref{sec:flat.UCB} that this algorithm has a finite regret $O(2^D/\Delta)$ (where $\Delta=\min_{i, \Delta_i>0} \Delta_i$) with high probability.

In Section \ref{sec:smoothness}, we introduce a UCB-based algorithm, called {\em Bandit Algorithm for Smooth Trees}, which takes into account actual smoothness of the rewards for performing efficient ``cuts'' of sub-optimal branches based on concentration inequality.
We give a numerical experiment for the problem of optimizing a Lipschitz function given noisy observations.

Finally, in Section \ref{sec:growing.tree} we present and analyze a growing tree search, which builds incrementally the tree by expanding, at each iteration, the most promising node. This method is memory efficient and well adapted to search in large (possibly infinite) trees.

\paragraph{Additional notations:}
Let $\L$ denotes the set of leaves and $\S$ the set of sub-optimal leaves. For any node $i$, we write $\L(i)$ the set of leaves in the branch starting from node $i$. For any node $i$, we write $n_i$  the number of times node $i$ has been visited up to round $n$, and we define the cumulative rewards:
$$X_{i,n_i}=\frac1{n_i}\sum_{j\in \L(i)} n_j X_{j,n_j},$$
the cumulative expected rewards:
$$\bar X_{i,n_i}=\frac1{n_i}\sum_{j\in \L(i)} n_j \mu_j,$$
and the pseudo-regret:
$$\bar R_{i,n_i}=\sum_{j\in \L(i)} n_j (\mu_i-\mu_j).$$

\section{Lower regret bound for UCT}\label{sec:UCT}

The UCT algorithm introduced in \cite{Kocsis06} is believed to adapt automatically to the effective (and a priori unknown) smoothness of the tree: If the tree possesses an effective depth $d<D$ (i.e. if all leaves of a branch starting from a node of depth $d$ have the same value) then its regret will be equal to the regret of a tree of depth $d$. First, we notice that the bound (\ref{eq:uct}) is not a true upper confidence bound on the value $\mu_i$ of a node $i$ since the rewards received at node $i$ are not identically distributed (because the chosen leaves depend on a non-stationary node selection process). However, due to the increasing confidence term $\log(p)$ when a node is not chosen, all nodes will be infinitely visited, which guarantees an asymptotic regret of $O(\log(n))$. However the transitory phase may last very long.

\begin{figure}[tbph]
\centerline{\epsfig{figure=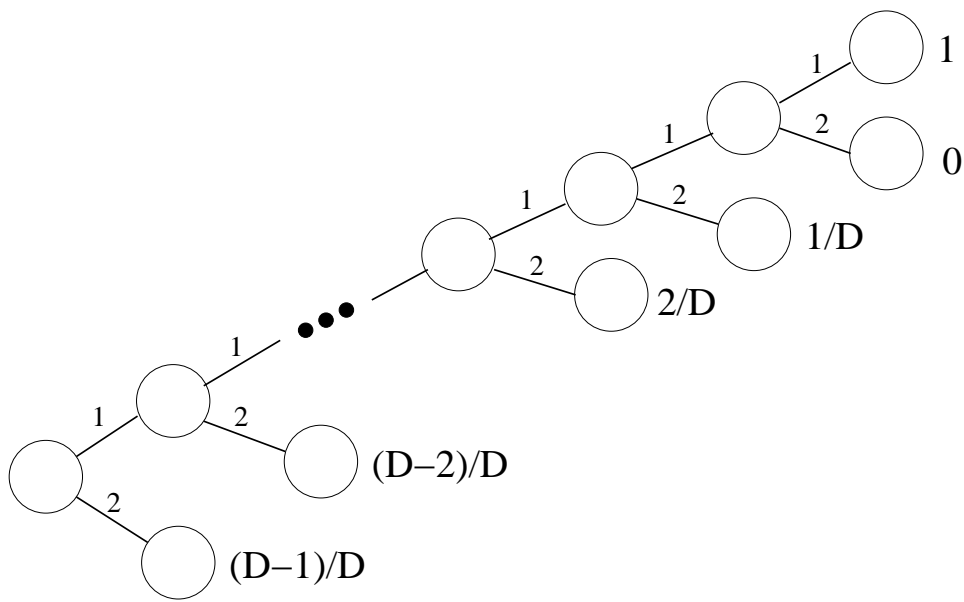,width=0.9\textwidth}}
\caption{A bad example for UCT. From the root (left node), action $2$ leads to a node from which all leaves yield reward $\frac{D-1}{D}$. The optimal branch consists in choosing always action $1$, which yields reward $1$. In the beginning, the algorithm believes the arm $2$ is the best, spending most of its times exploring this branch (as well as all other sub-optimal branches). It takes $\Omega(\exp(\exp(D)))$ rounds to get the $1$ reward!}
\label{fig:bad_tree}
\end{figure}

Indeed, consider the example illustrated in Figure \ref{fig:bad_tree}. The rewards are deterministic and for a node of depth $d$ in the optimal branch (obtained after choosing $d$ times action $1$), if action $2$ is chosen, then a reward of $\frac{D-d}{D}$ is received (all leaves in this branch have the same reward). If action $1$ is chosen, then this moves to the next node in the optimal branch. At depth $D-1$, action $1$ yields reward $1$ and action $2$, reward $0$. We assume that when a node is visited for the first time, the algorithm starts by choosing action $2$ before choosing action $1$.

We now establish a lower bound on the number of times suboptimal rewards are received before getting the optimal $1$ reward for the first time.
Write $n$ the first instant when the optimal leaf is reached. Write $n_d$ the number of times the node (also written $d$ making a slight abuse of notation) of depth $d$ in the optimal branch is reached. Thus $n=n_0$ and $n_D=1$. At depth $D-1$, we have $n_{D-1}=2$ (since action $2$ has been chosen once in node $D-1$).

We consider both the logarithmic confidence sequence used in (\ref{eq:uct}) and the square root sequence in (\ref{eq:uct_sqrt}). Let us start with the square root confidence sequence (\ref{eq:uct_sqrt}). 
At depth $d-1$, since the optimal branch is followed by the $n$-th trajectory, we have (writting $d'$ the node resulting from action $2$ in the node $d-1$):
$$X_{d',n_{d'}} + \sqrt{\frac{\sqrt{n_{d-1}}}{n_{d'}}} \leq X_{d,n_{d}} + \sqrt{\frac{\sqrt{n_{d-1}}}{n_{d}}}.$$

 But $X_{d',n_{d'}}=(D-d)/D$ and $X_{d,n_{d}}\leq(D-(d+1))/D$ since the $1$ reward has not been received before. We deduce that 
$$\frac{1}{D} \leq \sqrt{\frac{\sqrt{n_{d-1}}}{n_{d}}}.$$

Thus for the square root confidence sequence, we have $n_{d-1}\geq n_d^2 / D^4$.
Now, by induction,
$$ n\geq \frac{n_1^2}{D^4}\geq \frac{n_2^{2^2}}{D^{4(1+2)}}\geq \frac{n_3^{2^3}}{D^{4(1+2+3)}} 
\geq \cdots \geq \frac{n_{D-1}^{2^{D-1}}}{D^{2D(D-1)}}$$

Since $n_{D-1}=2$, we obtain $n\geq \frac{2^{2^{D-1}}}{D^{2D(D-1)}}$. This is a double exponential dependency w.r.t. $D$. For example, for  $D=20$, we have $n\geq 10^{156837}$. Consequently, the regret is also $\Omega(\exp(\exp(D)))$. 

Now, the usual logarithmic confidence sequence defined by (\ref{eq:uct}) yields an even worst lower bound on the regret since we may show similarly that $n_{d-1}\geq \exp(n_d /(2 D^2))$ thus $n\geq \exp(\exp(\cdots\exp(2)\cdots))$ (composition of $D-1$ exponential functions).

Thus, although UCT algorithm has asymptotically regret $O(\log(n))$ in $n$, (or $O(\sqrt{n})$ for the square root sequence), the transitory regret is  $\Omega(\exp(\exp(\cdots\exp(2)\cdots)))$ (or $\Omega(\exp(\exp(D)))$ in the square root sequence).

The reason for this bad behavior is that the algorithm is too optimistic (it does not explore enough and may take a very long time to discover good branches that looked initially bad) since the bounds (\ref{eq:uct}) and (\ref{eq:uct_sqrt}) are not true upper bounds.

\section{Modified UCT} \label{sec:modified.UCT}
We modify the confidence sequence to explore more the nodes close to the root that the leaves, taking into account the fact that the time needed to decrease the bias ($\mu_i - \E[X_{i,n_i}]$) at a node $i$ of depth $d$ increases with the depth horizon ($D-d$). For such a node $i$ of depth $d$, we define the upper confidence bound:
\beq\label{eq:modified-uct}
B_{i,n_i}\eqdef X_{i,n_i}+ (k_d+1) \sqrt{\frac{2\log(\beta_{n_i}^{-1})}{n_i}} + \frac{k'_d}{n_i},
\eeq
where $\beta_{n}\eqdef \frac{\beta}{2 N n (n+1)}$ with $N=2^{D+1}-1$ the number of nodes in the tree, and the coefficients:
\beqa\label{eq:kd}
k_d &\eqdef& \frac{1+\sqrt{2}}{\sqrt{2}} \big[ (1+\sqrt{2})^{D-d}-1\big] \\
k'_d &\eqdef& (3^{D-d}-1)/2 \notag
\eeqa
Notice that we used a simplified notation, writing $B_{i,n_i}$ instead of $B_{i,p,n_i}$ since the bound does not depend on  the number of visits of the parent's node.
\begin{theorem}\label{thm:modified-uct}
Let $\beta>0$. Consider Algorithm \ref{alg:BATS} with the upper confidence bound (\ref{eq:modified-uct}). Then, with probability at least $1-\beta$, for all $n\geq 1$, the pseudo-regret is bounded by
\beq \label{eq:regret.modif-uct} 
\bar R_n\leq \frac{1+\sqrt{2}}{\sqrt{2}}\big[(1+\sqrt{2})^D-1\big] \sqrt{2\log(\beta_{n}^{-1})n} + \frac{3^D-1}{2} \notag
\eeq
\end{theorem}

\proof We first remind Azuma's inequality (see \cite{Gyorfi96}):

\begin{prop}\label{prop:Azuma}
Let $f$ be a Lipschitz function of $n$ independent
random variables such that
$f-\E f=\sum_{i=1}^n d_i$ where $(d_i)_{1\leq i\leq n}$ is a
martingale difference sequence, i.e.: $\E[d_i|\FF_{i-1}]=0$, $1\leq
i\leq n$, such that $||d_i||_\infty\leq 1/n$. Then for every
$\epsilon>0$,
$$ \PP{|f-\E f|>\epsilon} \leq 2 \exp\big( - n \epsilon^2/2).$$
\end{prop}

We apply Azuma's inequality to the random variables $Y_{i,n_i}$ and $Z_{i,n_i}$, defined respectively, for all nodes $i$, by $Y_{i,n_i} \eqdef X_{i,n_i}-\bar X_{i,n_i}$, and for all non-leaf nodes $i$, by
$$Z_{i,n_i} \eqdef \frac{1}{n_i}(n_{i_1}Y_{i_1,n_{i_1}}-n_{i_2}Y_{i_2,n_{i_2}}),$$ 
where $i_1$ and $i_2$ denote the children of $i$.

Since at each round $t\leq n_i$, the choice of the
next leaf only depends on the random samples drawn at previous times $s<t$, we have $\E[Y_{i,t}|\F_{t-1}]=0$ and $\E[Z_{i,t}|\F_{t-1}]=0$,  (i.e.. $Y$ and $Z$ are martingale difference sequences), and Azuma's inequality gives that, for any node $i$, for any $n_i$, for any $\epsilon>0$, 
$\PP{|Y_{i,n_i}|>\epsilon} \leq  2 \exp\big( - n_i \epsilon^2/2)$ and
$\PP{|Z_{i,n_i}|>\epsilon} \leq  2 \exp\big( - n_i \epsilon^2/2)$.

We now define a confidence level $c_{n_i}$ such that with probability at least $1-\beta$, the random variables $Y_{i,n_i}$ and $Z_{i,n_i}$ belong to their
confidence intervals for all nodes  and for all times. More precisely, let $\EE$ be the event under which, for all $n_i\geq 1$, for all nodes $i$, $|Y_{i,n_i}| \leq c_{n_i}$ and for all non-leaf nodes $i$, $|Z_{i,n_i}| \leq c_{n_i}$. Then, by defining
$$c_{n}\eqdef \sqrt{\frac{2\log(\beta_{n}^{-1})}{n}}, \mbox{ with } \beta_n\eqdef \frac{\beta}{2N n (n+1)},$$
the event $\EE$ holds with  probability at least $1-\beta$.

Indeed, from an union bound argument, there are at most $2N$ inequalities (for each node, one for $Y$, one for $Z$) of the form:

$$\PP{|Y_{i,n_i}|>c_{n_i}, \forall n_i\geq 1} \leq \sum_{n_i\geq 1}
\frac{\beta}{2 N n_i (n_i+1)} = \frac{\beta}{2 N}.$$

We now prove Theorem \ref{thm:modified-uct} by bounding the pseudo-regret under the event $\EE$. We show by induction that the pseudo-regret at any node $j$ of depth $d$ satisfies:
\beq \label{eq:hyp.regret}
\bar R_{j,n_j}\leq k_d n_j c_{n_j}+k_d',
\eeq 
This is obviously true for $d=D$, since the pseudo-regret is zero at the leaves. Now, let a node $i$ of depth $d-1$. Assume that the regret at the childen's nodes satisfies (\ref{eq:hyp.regret}) (for depth $d$). For simplicity, write $1$ the optimal child and $2$ the sub-optimal one. Write $$c_{n}^d=(k_d+1) c_{n}+k_d'/n$$ the confidence interval defined by the choice of the bound (\ref{eq:modified-uct}) at a node of depth $d$.

If at round $n$, the node $2$ is chosen, this means that $X_{1,n_{1}} + c^d_{n_{1}}\leq X_{2,n_{2}} + c^d_{n_{2}}$. Now, since $|Z_{i,n_i}|\leq c_{n_i}$, we have $n_2(X_{2,n_2}-\bar X_{2,n_2})\leq n_1 Y_{1,n_1}+ n_i c_{n_i}$, thus:
$$ n_2 (X_{1,n_{1}} + c^d_{n_{1}}) \leq n_1 Y_{1,n_{1}} + n_{2} (\bar X_{2,n_{2}} + c^d_{n_{2}}) + n_i c_{n_i}.$$
Now, since, $|Y_{i,n_i}|\leq c_{n_i}$, we deduce that:
$$ n_2 (\bar X_{1,n_{1}} - \bar X_{2,n_{2}} - c_{n_1} + c^d_{n_{1}}) \leq n_1 c_{n_1} + n_2 c^d_{n_{2}} + n_i c_{n_i}.$$

Now, from the definitions of $\bar X$ and $\bar R$, we have: 
$$\bar X_{1,n_{1}} - \bar X_{2,n_{2}} = \mu_1 - \frac{\bar R_{1,n_{1}}}{n_1} - \big(\mu_2 - \frac{\bar R_{2,n_{2}}}{n_2}\big) = \Delta_i - \frac{\bar R_{1,n_{1}}}{n_1} + \frac{\bar R_{2,n_{2}}}{n_2},$$
where $\Delta_i\eqdef \mu_1-\mu_2$.
Thus, if action $2$ is chosen, we have:
$$n_2 (\Delta_i - \frac{\bar R_{1,n_1}}{n_1}+\frac{\bar R_{2,n_2}}{n_2} - c_{n_1}+c^d_{n_1})\leq n_1 c_{n_1}+n_i c_{n_i}+n_2 c^d_{n_2}.$$
From the definition of $c^d_{n_1}$ and the assumption (\ref{eq:hyp.regret}) on $\bar R_{1,n_1}$, we have $ - \bar R_{1,n_1}/n_1 - c_{n_1}+c^d_{n_1}\geq 0$, thus
\beqan 
n_2&\leq& \big[ n_1 c_{n_1}+n_i c_{n_i}+n_2 (k_d +1)c_{n_2} + k'_d\big]/\Delta_i \\
&\leq & \big[(1+\sqrt{2}+k_d)n_i c_{n_i} + k'_d\big]/\Delta_i.
\eeqan

Thus if $n_2> [(1+\sqrt{2}+k_d)n_i c_{n_i}+k'_d] /\Delta_i$, the arm $2$ will never be chosen any more. We deduce that for all $n\geq 0$,
$n_2\leq [(1+\sqrt{2}+k_d)n_i c_{n_i}+k'_d] /\Delta_i+1.$

Now, the pseudo-regret at node $i$ satisfies:
\beqan
\bar R_{i,n_i} &\leq& \bar R_{1,n_1}+\bar R_{2,n_2}+ n_2 \Delta_i \\
&\leq & (1+\sqrt{2})(1+k_d) n_i c_{n_i} + 3k'_d+\Delta_i,
\eeqan
which is of the same form as (\ref{eq:hyp.regret}) with 
\beqan
k_{d-1} &=& (1+\sqrt{2})(1+k_d) \\
k'_{d-1}&=& 3k'_d+1.
\eeqan

Now, by induction, given that $k_D=0$ and $k'_D=0$, we deduce the general form (\ref{eq:kd}) of $k_d$ and $k_d'$ and the bound on the pseudo-regret at the root for $d=0$.
\endproof

Notice that the $B$ values defined by (\ref{eq:modified-uct}) are true upper bounds on the nodes value: under the event $\EE$, for all node $i$, for all $n_i\geq 1$, $\mu_i\leq B_{i,n_i}$. Thus this procedure is safe, which prevents from having bad behaviors for which the regret could be disastrous, like in regular UCT. However, contrarily to regular UCT, in good cases, the procedure does not adapt to the effective smoothness in the tree. For example, at the root level, the confidence sequence is $O(\exp(D)/\sqrt{n})$ which lead to almost uniform sampling of both actions during a time $O(\exp(D))$. Thus, if the tree were to contain 2 branches, one only with zeros, one only with ones, this smoothness would not be taken into account, and the regret would be comparable to the worst-case regret.
Modified UCT is less optimistic than regular UCT but safer in a worst-case scenario.

\section{Flat UCB} \label{sec:flat.UCB}
A method that would combine both the safety of modified UCT and the adaptivity of regular UCT is to consider a regular UCB algorithm on the leaves. Such a {\em flat UCB} could naturally be implemented in the tree structure by defining the upper confidence bound of a non-leaf node as the maximal value of the children's bound:  
\beq\label{eq:flat-ucb}
B_{i,n_i} \eqdef
\left\{
\begin{array}{ll}
X_{i,n_i} + \sqrt{\frac{2\log(\beta_{n_i}^{-1})}{n_i}} & \mbox{ if } i \mbox{ is a leaf,} \\
\max\big[ B_{i_1,n_{i1}}, B_{i_2,n_{i2}}\big] & \mbox{ otherwise.}
\end{array}
\right.
\eeq
where we use $$\beta_{n}\eqdef\frac{\beta}{2^D n (n+1)}.$$

We deduce:

\begin{theorem} \label{thm:flat-ucb} Consider the flat UCB defined by Algorithm \ref{alg:BATS} and (\ref{eq:flat-ucb}).
Then, with probability at least $1-\beta$, the pseudo-regret is bounded by a constant:
$$ \bar R_n \leq 40 \sum_{i\in \S} \frac{1}{\Delta_i}\log(\frac{2^{D+1}}{\Delta_i^2\beta}) 
\leq 40 \frac{2^D}{\Delta}\log(\frac{2^{D+1}}{\Delta^2\beta}),$$
where $\S$ is the set of sub-optimal leaves, i.e.. $\S=\{ i\in \L, \Delta_i>0\}$, and $\Delta = \min_{i\in \S} \Delta_i$. 
\end{theorem}

\proof Consider the event $\EE$ under which, for all leaves $i$, for all $n\geq 1$, we have $|X_{i,n}-\mu_i|\leq c_{n}$, with the confidence interval $c_n=\sqrt{\frac{2\log(\beta_{n}^{-1})}{n}}$. Then, the event $\EE$ holds with probability at least $1-\beta$. Indeed, as before, using an union bound argument, there are at most $2^D$ inequalities (one for each leaf) of the form:

$$\PP{|X_{i,n}-\mu_i|>c_{n}, \forall n\geq 1} \leq \sum_{n\geq 1}
\frac{\beta}{2^D n (n+1)} = \frac{\beta}{2^D}.$$

Under the event $\EE$, we now provide a regret bound by bounding the number of times each sub-optimal leaf is visited. Let $i\in\S$ be a sub-optimal leaf. Write $*$ an optimal leaf. If at some round $n$, the leaf $i$ is chosen, this means that 
$X_{*,n_{*}} + c_{n_*} \leq X_{i,n_{i}} + c_{n_i} $. Using the (lower and upper) confidence interval bounds for leaves $i$ and $*$, we deduce  that
$\mu^* \leq \mu_i + 2c_{n_i}$. Thus
$\big(\frac{\Delta_i}{2}\big)^2 \leq  \frac{2\log(\beta_{n_i}^{-1})}{n_i}.$
Hence, for all $n\geq 1$, $n_i$ is bounded by the smallest integer $m$ such that $\frac{m}{\log(\beta_m^{-1})} > 8/\Delta_i^2$. Thus $\frac{n_i-1}{\log(2^D n_i(n_i-1) \beta^{-1})} \leq w$, writing $w=8/\Delta_i^2$. This implies
\beq \label{eq:no-name}
n_i \leq 1+w\log(2^D n_i^2 \beta^{-1})
\eeq
A first rough bound yields $n_i\leq w^2 2^{D-2} \beta^{-1}$, which can be used to derive a tighter upper bound on $n_i$. After two recursive uses of (\ref{eq:no-name}) we obtain:
$$ n_i \leq 5w\log(w 2^{D-2} \beta^{-1}).$$
Thus, for all $n\geq 1$, the number of times leaf $i$ is chosen is at most $40 \log(2^{D+1}\beta^{-1}/\Delta_i^2)/\Delta_i^2$.
The bound on the regret follows immediately from the property that $\bar R_n= \sum_{i\in \S} n_i \Delta_i$.
\endproof

This algorithm is safe in the same sense as previously define, i.e. with high probability, the bounds defined by (\ref{eq:flat-ucb}) are true upper bounds on the value on the leaves. However, since there are $2^D$ leaves, the regret still depends exponentially on the depth $D$. 

\begin{remark} Modified UCT has a regret $O(2^D\sqrt{n})$ whereas Flat UCB has a regret $O(2^D/\Delta)$. The non dependency w.r.t. $\Delta$ in Modified UCT, obtained at a price of an additional $\sqrt{n}$ factor, comes from the application of Azuma's inequality also to $Z$, i.e. the difference between the children's deviations $X-\bar X$. An similar analysis in Flat UCB would yield a regret $O(2^D \sqrt{n})$.
\end{remark}

In the next section, we consider another UCB-based algorithm that takes into account possible smoothness of the rewards to process effective ``cuts'' of sub-optimal branches with high confidence.

\section{Bandit Algorithm for Smooth Trees} \label{sec:smoothness}
We want to exploit the fact that if the leaves of a branch have similar values, then a confidence interval on that branch may be made much tighter than the maximal confidence interval of its leaves (as processed in the Flat UCB). Indeed, assume that from a node $i$, all leaves $j\in\L(i)$ in the branch $i$ have values $\mu_j$, such that $\mu_i - \mu_j\leq \delta$. Then, 
$$ \mu_i \leq \frac{1}{n_i}\sum_{j\in \L(i)} n_j(\mu_{j}+\delta)\leq X_{i,n_i} +\delta + \bar X_{i,n_i}-X_{i,n_i},$$
and thanks to Azuma's inequality, the term $\bar X_{i,n_i}-X_{i,n_i}$ is bounded with probability $1-\beta$ by a confidence interval $\sqrt{\frac{2\log(\beta^{-1})}{n_i}}$ which depends only on $n_i$ (and not on $n_j$ for $j\in\L(i)$).
We now make the following assumption on the rewards:

\paragraph{Smoothness assumption:} Assume that for all depth $d<D$, there exists $\delta_d>0$, such that for any node $i$ of depth $d$, for all leaves $j\in\L(i)$ in the branch $i$, we have $\mu_i - \mu_j \leq \delta_d$.

Typical choices of the smoothness coefficients $\delta_d$ are exponential $\delta_d\eqdef\delta \gamma^d$ (with $\delta>0$ and $\gamma<1$), polynomial $\delta_d\eqdef \delta d^\alpha$ (with $\alpha<0$), or linear $\delta_d\eqdef \delta (D-d)$ (Lipschitz in the tree distance) sequences.

We define the {\em Bandit Algorithm for Smooth Trees} (BAST) by Algorithm \ref{alg:BATS} with the upper confidence bounds defined, for any leaf $i$, by
$B_{i,n_i} \eqdef X_{i,n_i} + c_{n_i},$ and for any non-leaf node $i$ of depth $d$, by
\beq \label{eq:BAST}
B_{i,n_i} \eqdef \min \Big\{ \max\big[ B_{i_1,n_{i1}}, B_{i_2,n_{i2}}\big], X_{i,n_i} + \delta_d+ c_{n_i} \Big\}
\eeq
with the confidence interval $$c_n\eqdef\sqrt{\frac{2\log(N n (n+1) \beta^{-1})}{n}}.$$

We now provide high confidence bounds on the number of times each sub-optimal node is visited.
\begin{theorem}\label{thm:BAST1}
Let $I$ denotes the set of nodes i such that $\Delta_i>\delta_{d_i}$, where $d_i$ is the depth of node $i$. Define recursively the values $N_i$ associated to each node $i$ of a sub-optimal branch (i.e. for which $\Delta_i>0$):

-  If $i$ is a leaf, then $$N_i \eqdef\frac{40\log(2 N \beta^{-1} /\Delta_i^2)}{\Delta_i^2},$$

-  It $i$ is not a leaf, then
$$
N_i \eqdef \left\{ 
\begin{array}{ll}
\hspace{-2mm} N_{i_1}+N_{i_2}, & \hspace{-4mm}\mbox{ if } i\notin I \\
\hspace{-2mm} \min(N_{i_1}+N_{i_2}, \frac{40\log(2N \beta^{-1}/(\Delta_i-\delta_{d_i})^2)}{(\Delta_i-\delta_{d_i})^2} ), & \hspace{-4mm}\mbox{ if } i\in I
\end{array}
\right.
$$
where $i_1$ and $i_2$ are the children nodes of $i$. Then, with probability $1-\beta$, for all $n\geq 1$, for all sub-optimal nodes $i$, $n_i\leq N_i$.
\end{theorem}

\proof We consider the event $\EE$ under which $|X_{i,n}-\bar X_{i,n}|\leq c_n$ for all nodes $i$ and all times $n\geq 1$. The confidence interval $c_n=\sqrt{\frac{2\log(N n (n+1) \beta^{-1})}{n}}$ is chosen such that $\PP{\EE}\geq 1-\beta$. Under $\EE$, using the same analysis as in the Flat UCB, we deduce the bound  $n_i\leq N_i$ for any sub-optimal leaf $i$. 

Now, by backward induction on the depth, assume that $n_i\leq N_i$ for all sub-optimal nodes of depth $d+1$. Let $i$ be a node of depth $d$. Then $n_i \leq n_{i_1}+n_{i_2}\leq N_{i_1}+N_{i_2}$. 

Now consider a sub-optimal node $i\in I$. If the node $i$ is chosen at round $n$, the form of the bound (\ref{eq:BAST}) implies that for any optimal node $*$, we have $B_{*,n_*}\leq B_{i,n_i}$. Under $\EE$, $\mu^*\leq B_{*,n_*}$ and $B_{i,n_i}\leq X_{i,n_i}+\delta_d+c_{n_i}\leq \mu_i +\delta_d+2 c_{n_i}$. Thus $\mu^* \leq \mu_i +\delta_d + 2c_{n_i}$, which rewrites $\Delta_i-\delta_d \leq 2c_{n_i}$. Using the same argument as in the proof of Flat UCB, we deduce that for all $n\geq 1$, we have $n_i \leq \frac{40\log(2N \beta^{-1}/(\Delta_i-\delta_{d_i})^2)}{(\Delta_i-\delta_{d_i})^2}$. Thus $n_i\leq N_i$ at depth $d$, which finishes the inductive proof. \endproof
  
Now we would like to compare the regret of BAST to that of Flat UCB. First, we expect a direct gain for nodes $i\in I$. Indeed, from the previous result, whenever a node $i$ of depth $d$ is such that $\Delta_i>\delta_d$, then this node will be visited, with high probability, at most $O(1/(\Delta_i-\delta_d)^2)$ times (neglecting $\log$ factors). But we also expect an upper bound on $n_i$ whenever $\Delta_i>0$ if at a certain depth $h\in [d,D]$, all nodes $j$ of depth $h$ in the branch $i$ satisfy $\Delta_j>\delta_h$. 

The next result enables to further analyze the expected improvement over Flat UCB.

\begin{theorem}
Consider the exponential assumption on the smoothness coefficients : $\delta_d \leq \delta \gamma^d$. For any $\eta\geq 0$ define the set of leaves $I_\eta\eqdef\{i\in \S, \Delta_i\leq \eta\}$. Then, with probability at least $1-\beta$, the pseudo regret satisfies, for all $n\geq 1$,
\beqa\label{eq:bound.regret}
\bar R_n &\leq& \sum_{i\in I_\eta, \Delta_i>0} \frac{40}{\Delta_i}\log(\frac{2N\beta^{-1}}{\Delta_i^2}) +
|I_\eta|\frac{320(2\delta)^c}{\eta^{2+c}} \log(\frac{2N}{\eta^2\beta}) \notag \\ 
&\leq & 40|I_\eta| \Big( \frac{1}{\Delta}\log(  \frac{2N}{\Delta^2\beta}) +
\frac{8(2\delta)^c}{\eta^{2+c}} \log(\frac{2N}{\eta^2\beta}) \Big)
\eeqa
where $$c\eqdef \log(2)/\log(1/\gamma).$$
\end{theorem}

 Note that this bound (\ref{eq:bound.regret}) does not depend explicitly on the depth $D$. Thus we expect this method to scale nicely in big trees (large $D$). The first term in the bound is the same as in Flat UCB, but the sum is performed only on leaves $i\in I_\eta$ whose value is $\eta$-close to optimality. Thus, BAST is expected to improve over Flat UCB (at least as expressed by the bounds) whenever the number $|I_\eta|$ of $\eta$-optimal leaves is small compared to the total number of leaves $2^D$. In particular, with $\eta<\Delta$, $|I_\eta|$ equals the number of optimal leaves, so taking $\eta\rightarrow\Delta$ we deduce a regret $O(1/\Delta^{2+c})$.

\proof We consider the same event $\EE$ as in the proof of Theorem \ref{thm:BAST1}.
Call ``$\eta$-optimal'' a branch that contains at least a leaf in $I_\eta$. Let $i$ be a node, of depth $d$, that does not belong to an $\eta$-optimal branch. Let $h$ be the smallest integer such that $\delta_h\leq \eta/2$, where $\delta_h=\delta \gamma^h$. We have $h \leq \frac{\log(2\delta/\eta)}{\log(1/\gamma)}+1$. Let $j$ be a node of depth $h$ in the branch $i$. Using similar arguments as in Theorem \ref{thm:flat-ucb}, the number of times $n_j$ the node $j$ is visited is at most $$n_j\leq  \frac{40\log(2N\beta^{-1}/(\Delta_j-\delta_h)^2)}{(\Delta_j-\delta_h)^2},$$ but since $\Delta_j-\delta_h\geq \Delta_i - \delta_h\geq \Delta_i - \eta/2 \geq \eta/2$, we have:
$ n_j\leq l/\eta^2$, writing $l=160\log(8N\beta^{-1}/\eta^2).$

Now the number of such nodes $j$ is at most $2^{h-d}$, thus:
\beqan 
n_i&\leq& 2^{h-d} \frac{l}{\eta^2}\leq 2 \Big( \frac{2\delta}{\eta} \Big)^{\frac{\log(2)}{\log(1/\gamma)}} 2^{-d}\frac{l}{\eta^2} 
= \frac{2 l (2\delta)^c}{\eta^{c+2}} 2^{-d}
\eeqan
with $c=\log(2)/\log(1/\gamma)$. Thus, the number of times $\eta$-optimal branches are not followed until the $\eta$-optimal leaves is at most 
$$|I_\eta| \sum_{d=1}^D \frac{2 l (2\delta)^c}{\eta^{c+2}} 2^{-d}\leq |I_\eta|\frac{2 l (2\delta)^c}{\eta^{c+2}}.$$

Now for the leaves $i\in I_\eta$, we derive similarly to the Flat UCB that $n_i\leq 40\log(2N\beta^{-1}/\Delta_i^2)/\Delta_i^2$. Thus, the pseudo regret is bounded by the sum for all sub-optimal leaves $i\in I_\eta$ of $n_i \Delta_i$ plus the sum of all trajectories that do not follow $\eta$-optimal branches until $\eta$-optimal leaves $|I_\eta|\frac{2 l (2\delta)^c}{\eta^{c+2}}$. This implies (\ref{eq:bound.regret}).
\endproof

\begin{remark}
Notice that if we choose $\delta=0$, then BAST algorithm reduces to regular UCT (with a slightly different confidence interval), whereas if $\delta=\infty$, then this is simply Flat UCB. Thus BAST may be seen as a generic UCB-based bandit algorithm for tree search, that allows to take into account actual smoothness of the tree, if available. 
\end{remark}

\paragraph{Numerical experiments: global optimization of a noisy function.}
We search the global optimum of an $[0,1]$-valued function, given noisy data. The domain $[0,1]$ is uniformly discretized by $2^D$ points $\{y_j\}$, each one related to a leaf $j$ of a tree of depth $D$. The tree implements a recursive binary splitting of the domain. At time $t$, if the algorithm selects a leaf $j$, then the (binary) reward $x_t\stackrel{i.i.d.}{\sim} \B(f(y_j))$, a Bernoulli random variable with parameter $f(y_j)$ (i.e. $\PP{x_t=1}=f(y_j)$, $\PP{x_t=0}=1-f(y_j)$). 

\begin{figure}[htbp]
\includegraphics[width=12cm,height=7cm]{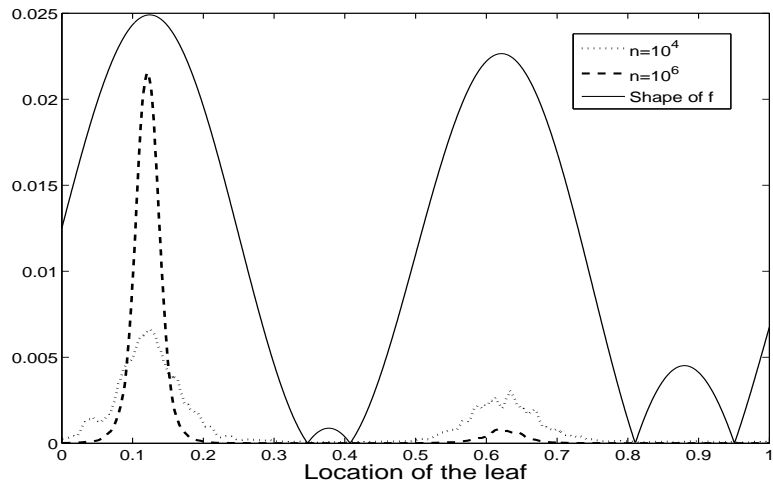}
\caption{Function $f$ rescaled (in plain) and proportion $n_j/n$ of leaves visitation for BAST with $\delta=7$, depth $D=10$, after $n=10^4$  and $n=10^6$ rounds.}
\label{fig:hitogramN}
\end{figure}

We assume that $f$ is Lipschitz. Thus the exponential smoothness assumption  $\delta_d=\delta 2^{-d}$ on the rewards holds with $\delta$ being the Lipschitz constant of $f$ and $\gamma=1/2$ (thus $c=1$).
In the experiments, we used the function 
$$f(x) \eqdef |\sin(4\pi x)+\cos(x)|/2$$ 
plotted in Figure \ref{fig:hitogramN}. Note that an immediate upper bound on the Lipschitz constant of $f$ is $(4\pi+1)/2<7$.

\begin{center}
\begin{figure}[hbpt]
\includegraphics[width=12cm,height=6cm]{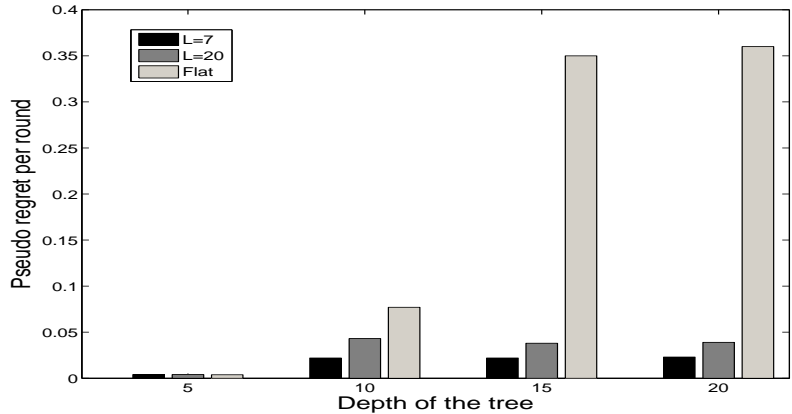}
\caption{Pseudo regret per round $\bar R_n/n$ for $n=10^6$, as a function of the depth $D\in\{5,10,15,20\}$, for BAST with $\delta\in\{7,20\}$ and Flat UCB.}
\label{fig:RNetRD}
\end{figure}
\end{center}

We compare Flat UCB and BAST algorithm for different values of $\delta$. 
Figure \ref{fig:RNetRD} show the respective pseudo-regret per round $\bar R_n/n$ for BAST used with a good evaluation of the Lipschitz constant ($\delta=7$), BAST used with a poor evaluation of the Lipschitz constant ($\delta=20$), and Flat UCB ($\delta=\infty$).
As expected, we observe that BAST outerforms Flat UCB, and that the performance of BAST is less dependent of the size of the tree than Flat UCB. BAST with a poor evaluation of $\delta$ still performs better than Flat UCB.
BAST concentrates its ressources on the good leaves: In Figure \ref{fig:hitogramN} we show the proportion $n_j/n$ of visits of each leaf $j$. We observe that, when $n$ increases, the proportion of visits concentrates on the leaves with highest $f$ value.

\begin{remark}
 If we know in advance that the function is smooth (e.g. of class $\C^2$ with bounded second order derivative), then one could use Taylor's expansion to derive much tighter upper bounds, which would cut more efficiently sub-optimal branches and yield improved performance. Thus any a priori knowledge about the tree smoothness could be taken into account in the BAST bound (\ref{eq:BAST}).
\end{remark}

\section{Growing trees} \label{sec:growing.tree}
If the tree is too big (possibly infinite) to be represented, one may wish to discover it iteratively, exploring it as the same time as searching for an optimal value. We propose an incremental algorithm similar to the method described in \cite{Coulom06} and \cite{Gelly06}: The algorithm starts with only the root node. Then, at each stage $n$ it chooses which leaf, call it $i$, to expand next. Expanding a leaf means turning it into a node $i$, and adding in our current tree representation its children leaves $i_1$ and $i_2$, from which a reward (one for each child) is received. The process is then repeated in the new tree.

We make an assumption on the rewards: from any leaf $j$, the received reward $x$ is a random variable whose expected value satisfies: $ \mu_j - \E[x] \leq \delta_d$, where $d$ is the depth of $j$, and $\mu_j$ the value of $j$ (defined as previously by the maximum of its children values).

Such an iterative growing tree requires a amount of memory $O(n)$ directly related to the number of exploration rounds in the tree. 

We now apply BAST algorithm and expect the tree to grow in an asymmetric way, expanding in depth the most promising branches first, leaving mainly unbuilt the sub-optimal branches.

\begin{theorem} Consider this incremental tree method using Algorithm \ref{alg:BATS} defined by the bound (\ref{eq:BAST}) with the confidence interval 
$$c_{n_i}\eqdef\sqrt{\frac{2\log(n (n+1) n_i (n_i+1) \beta^{-1})}{n_i}},$$ where $n$ is the current number of expanded nodes. Then, with probability $1-\beta$, for any sub-optimal node $i$ (of depth $d$), i.e. s.t. $\Delta_i>0$, 
\beq \label{eq:BAST2}
n_i \leq 10 \Big(\frac{\delta}{c}\Big)^c \Big(\frac{2+c}{\Delta_i}\Big)^{c+2}  \log\Big(\frac{n(n+1)}{\beta} \frac{(2+c)^2}{2\Delta_i^2}\Big)2^{-d}.
\eeq
\end{theorem}

Thus this algorithm essentially develops the optimal branch, i.e. except for $O(\log(n))$ samples at each depth, all computational resources are devoted to further explore the optimal branch. 

\proof Consider the event $\EE$ under which $|X_{i,n_i}-\bar X_{i,n_i}|\leq c_{n_i}$ for all expanded nodes $1\leq i\leq n$, all times $n_i\geq 1$, and all rounds $n\geq 1$. The confidence interval $c_{n_i}$ are such that $\PP{\EE}\geq 1-\beta$. At round $n$, let $i$ be a node of depth $d$. Let $h$ be a depth such that $\delta_h< \Delta_i$. This is satisfied for all integer $h\geq \frac{\log(\delta/\Delta_i)}{\log(1/\gamma)}$. Similarly to Flat UCB, we deduce that the number of times $n_j$ a node $j$ of depth $h$ has been visited is bounded by $40\log(2n(n+1)\beta^{-1}/(\Delta_i-\delta_h)^2)/(\Delta_i-\delta_h)^2$. Thus $i$ has been visited at most 
$$ n_i \leq \min_{h\geq \frac{\log(\delta/\Delta_i)}{\log(1/\gamma)}} 2^{h-d}
\frac{40\log(2n(n+1)\beta^{-1}/(\Delta_i-\delta_h)^2)}{(\Delta_i-\delta_h)^2}.$$
This function is minimized (neglecting the $\log$ term) for $h=\log(\frac{\delta(2+c)}{c\Delta_i})/\log(1/\gamma)$, which leads to (\ref{eq:BAST2}).
\endproof

\begin{figure}[tbph]
\includegraphics[width=12cm,height=10cm]{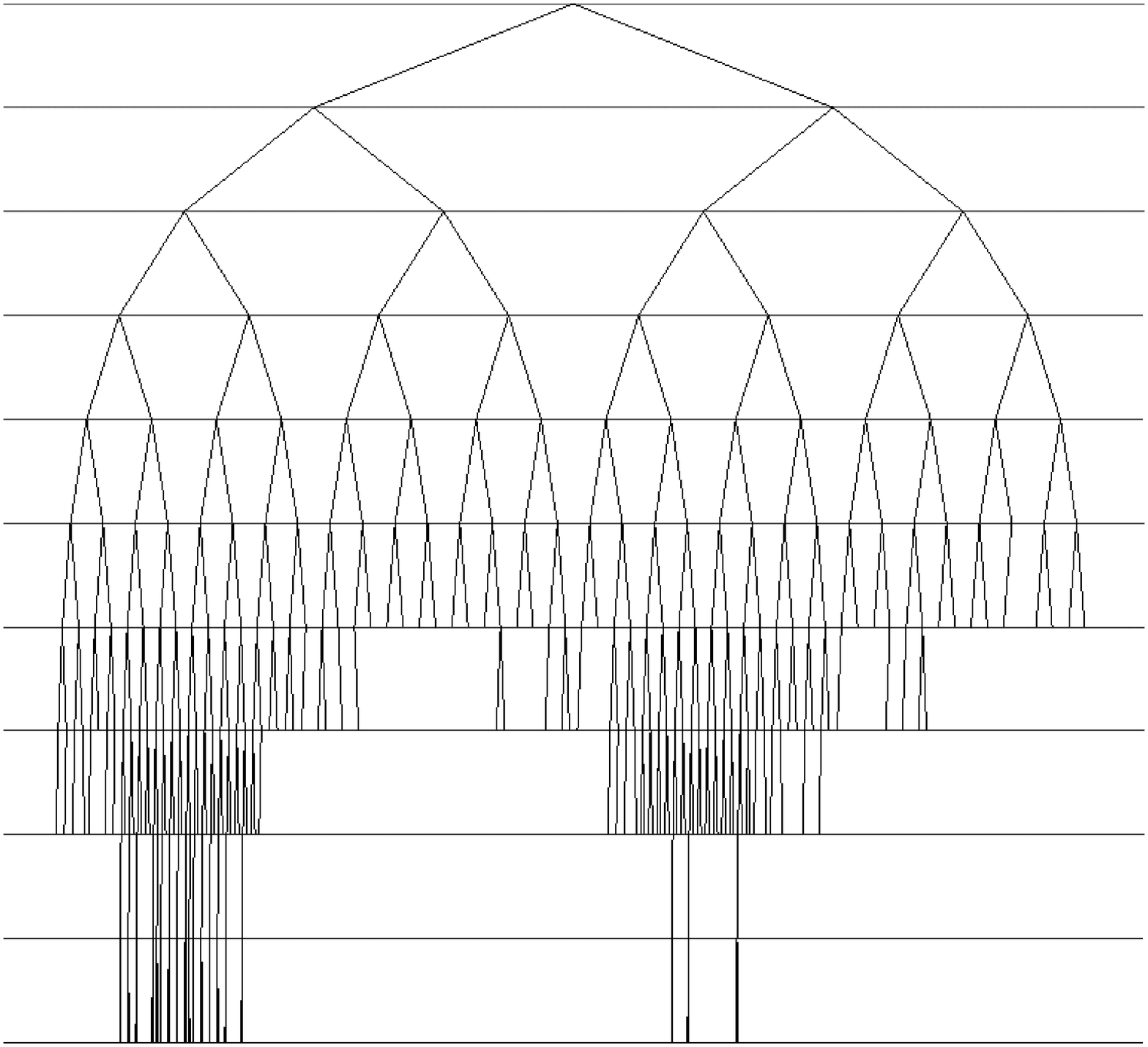}
\caption{Tree resulting from the iterative growing BAST algorithm, after $n=300$ rounds, with $\delta=7$.}
\label{fig:growingtree}
\end{figure}

For illustration, Figure \ref{fig:growingtree} shows the tree obtained applied to the function optimization problem of previous section, after $n=300$ rounds. The most in-depth explored branches are those with highest value.

\section{Conclusion}
We analyzed several UCB-based bandit algorithms for tree search. The good exploration exploitation tradeoff of these methods enables to return rapidly a good value, and improve precision if more time is provided. BAST enables to take into account possible smoothness in the tree to perform efficient ``cuts''\footnote{Note that this term may be misleading here since the UCB-based methods described here never explicitly delete branches} of sub-optimal branches, with high probability. 

If additional smoothness information is provided, the $\delta$ term in the bound (\ref{eq:BAST}) may be refined, leading to improved performance. Empirical information, such as variance estimate, could improve knowledge about local smoothness which may be very helpful to refine the bounds. However, it seems important to use true upper confidence bounds, in order to avoid bad cases as illustrated in regular UCT.
Applications include minimax search for games in large trees, and global optimization under uncertainty.

\subsection*{Acknowledgements} We wish to thank Jean-Fran\c cois Hren for running the numerical experiments.

\small
\bibliographystyle{alpha}
\bibliography{biblio}

\end{document}